\documentclass{AISB2008}
\usepackage{times}
\usepackage{graphicx}
\usepackage{latexsym}

\def\wvv{{\tt word2vec}} % TODO is this correct

\def\skipgr{Skip-gram}

\usepackage{tikz}
\usetikzlibrary{calc}
\usetikzlibrary{arrows}
\usetikzlibrary{shapes}
\usetikzlibrary{fit}

\usepackage{url}

\begin{document}

\title{Semantic Holism and Word Representations\\ in Artificial Neural Networks}

\author{Tom\'{a}\v{s} Musil\institute{%
Charles University, Faculty of Mathematics and Physics,
Institute of Formal and Applied Linguistics,
Malostranské náměstí 25, 118 00 Prague, Czech Republic, email: musil@ufal.mff.cuni.cz}}

\maketitle
\bibliographystyle{AISB2008}

\begin{abstract}
    Artificial neural networks are a state-of-the-art solution for many problems in natural language processing. What can we learn about language and meaning from the way artificial neural networks represent it?
    Word representations obtained from the \skipgr{} variant of the word2vec model exhibit interesting semantic properties.
	This is usually explained by referring to the general distributional hypothesis, which states that
	the meaning of the word is given by the contexts where it occurs. We propose a more
	specific approach based on Frege's holistic and functional approach to meaning. Taking
	Tugendhat's formal reinterpretation of Frege's work as a starting point, we demonstrate that it is analogical to the process of training the \skipgr{} model and offers a possible explanation
	of its semantic properties.
\end{abstract}

\section{INTRODUCTION} %lepší název?

“Meaning is, therefore, something that words have \emph{in sentences}; and it's something that sentences have \emph{in a language}.” \cite{fodor1992holism}
On the other hand, meaning could also be something that words have on their own, with sentences being compositions and language a collection of words. This is the question of semantic \emph{holism} versus \emph{atomism}, which was important in the philosophy of language in the second half of the 20th century and has not been satisfyingly answered yet.

Artificial neural networks are the state-of-the-art solution for many problems in natural language processing (and machine learning in general). They produce word representation with interesting properties, but the way they work is little understood from the perspective of linguistics or the philosophy of language.

We believe that by finding parallels between concepts in AI and the philosophy of language, we can better understand both areas.

In this paper, we present an analogy between meaning defined as \emph{truth-value potential} (a reformulation of Fregean holistic and functional approach) and a variant of language representation model, therefore pointing out a possibility that its “striking syntactic and semantic properties” \cite{mik} are formed due to adhering to holistic principles. 

\subsection{Related work}

We have found only one work concerning the philosophical aspects of neural language
models \cite{hon:pan}. It is, however, concentrating on Self-Organizing Maps
and Quine's version of semantic holism.

There are papers showing that \skipgr{}  with negative sampling is implicitly
a factorization of a word-context matrix (e.g. \cite{levy2014neural}, although this result was
later contested by various authors, such as \cite{arora2015} and
\cite{wilson2016}), or deriving the equations in an alternative way \cite{goldberg2014} (discussed more in  Section~\ref{ref:goldberg}). This may tell us something about the model, but it does not answer the principal question: why should the
 matrix factorized in a certain way contain semantic information?

\section{SEMANTIC HOLISM AND ATOMISM}

\emph{Semantic holism} (or \emph{meaning holism}) is “the thesis that what a linguistic expression
means depends on its relations to many or all other expressions within the same totality. [\ldots{}] The totality
in question may be the language to which the expressions belong, or a theory formulation
in that language.” \cite{lepore_meaning_2009}
The opposing view is called \emph{semantic atomism}, and it claims that there are expressions (typically words), whose meaning does not depend on the meaning of other expressions. The meaning of these expressions is given by something outside language (e.g. their relation to physical or mental objects).

In the following sections, we will specify the implications of both alternatives for semantics. The question also plays a role in cognitive science (content identity and similarity), epistemology (commensurability of theories) and seems to be strongly connected with the analytic/synthetic distinction \cite{fodor1992holism}. %TODO anything else?

There are other positions in between these two, such as \emph{semantic molecularism} or the belief that neither relations external nor internal are primary in forming meaning. However, to keep this text simple, we will only concentrate on extreme positions. We will also only talk about words,  although the same argument can be used with smaller meaningful language units (e.g. parts of a compound word).

Our goal is not to asses whether the truth lies with holism, atomism or neither of them. We will only show that holism is a useful perspective when understanding neural language models is concerned.

Before we get into details of the two perspectives, let us point out two critical aspects of their difference: holism proclaims \emph{interdependence} of meanings of words, contrary to their \emph{independence} in atomism. And holism favours \emph{decomposition} over \emph{composition}.

\subsection{Atomism}

“It is a widely held view that much of the history of the philosophy of language consists of a failed attempt to make semantic atomism work.” \cite[p. 32]{fodor1992holism}

Atomism played an important role in analytic philosophy, starting with Bertrand Russell's logical atomism and continuing with logical positivism, as exemplified in this quote by Carnap \cite{carnap1932}:
\begin{quote}
    A language consists of a vocabulary and a syntax, i.e. a set of words which have meanings and rules of sentence formation. These rules indicate how sentences may be formed out of the various sorts of words.
\end{quote}
For logical positivists, words have meaning, because they refer to objects (be it physical, sensual, logical, mathematical or other). The rules of composition determine the meaning of sentences (and rule out senseless sequences of words).

Under this (or similar) view, the fact that words refer to the outside world is presupposed. Their references are \emph{independent} of each other (that “dog” refers to \emph{dog} is independent of that “horse” refers to \emph{horse}).  There is strong emphasis on \emph{compositionality}, that reached its peak in Chomskian linguistics and is still relevant today.

Crucially, this means that a word can have meaning on its own (e.g. by referring to something). The meaning of larger units, such as sentences, is \emph{derived} by the rules of composition from the meaning of words.

\subsection{Holism}

Semantic holism accents the \emph{interdependence} of meaning. The \emph{whole} (language, theory, \ldots{}) is the primary vehicle of meaning. The meaning of smaller units is derived by \emph{decomposition}.

This view is motivated by the same word having a different meaning in a different context. Gottlob Frege has shown \cite{frege1884grundlagen} that even such seemingly unambiguous words as numbers play distinct roles in different situations: “5 is a prime number” and “there are 5 cows on the meadow” are different at least in that the first “5” signifies a complete (abstract) object, while the second one needs to be supplemented with information that it is cattle of which there are 5 specimens, otherwise the expression would not be grammatical.

Frege promoted what we could call \emph{sentence holism}:
“Only in the context of a sentence does a word have a meaning.” \cite{frege1892sinn} We will later use its modern reformulation to show an analogy with certain neural language models and therefore their holistic character. 

Another group of arguments for holism consist of variations on the theme of impossibility of knowing or using a word without being able to use other words. For example, it could be argued that a person could not correctly use the word “mammal”, without also knowing (at least some of) “bird”, “animal” and kinds of animals. Therefore the meaning of words cannot be formed in isolation. 

Something that is harder to explain under holism than under atomism is the fact that words refer to objects. If the meaning of words is given by other words, how is it connected to the world around us? However, not all words refer to something. And even if subscribing to holism makes explaining reference harder, it may be because it is a hard problem to explain.

Another thing that is simpler under atomism is compositionality. While in atomism it plays a central role as one of the presupposed properties of language, holism may not need it. But it does not claim that words \emph{do not have} meanining at all, only that it is \emph{derived} (by some sort of decomposition) from the meaning of the whole.

\section{WORD REPRESENTATIONS IN AI}

Although all artificial neural networks that work with language must have some way of representing it, the most interesting representations come from neural language models. Language modelling is a task of predicting a missing word from a sequence or generating text. There is also a similar class of models that are designed specifically to produce representations of language units, which we will call neural language representation models.

The representations (also called \emph{embeddings}) are high dimensional vectors of real numbers. They are either learned together with the rest of the network for the particular task or pretrained by a general language representation model (typically on a larger dataset not specific for the task).

Some neural language (representation) models produce
representation  with semantic properties, although the task of
language modeling itself is not (at least at the first sight) directly
connected with semantics and no explicit semantic annotation is given to the neural network. 

These semantic properties became popular with the invention of the \wvv{}
software and the \skipgr{} model, whose author said about it
\cite{mik}: 
\begin{quote}
The model itself has no knowledge of syntax or morphology or semantics.
Remarkably, training such a purely lexical model to maximize likelihood will
induce word representations with striking syntactic and semantic properties.
\end{quote}
%
%Although they find the properties ``striking'' and the fact that they exist ``remarkable'',
However, they did not present any explanation of the phenomenon.

\label{ref:goldberg}
Goldberg and Levy \cite{goldberg2014} present a detailed derivation of the
central equation of the \skipgr{} model. In the last section they say:
\begin{quote}

Why does this produce good word representations?

Good question. We don't really know.

The distributional hypothesis states that words in similar contexts have
similar meanings. The objective [of the \skipgr{} model] clearly tries to increase the
%quantity $v_w \cdot v_c$
[dot product of the context and the word representations]
for good word-context pairs, and decrease it for bad ones.
Intuitively, this means that words that share many contexts will be similar to
each other (note also that contexts sharing many words will also be similar to
each other). This is, however, very hand-wavy.  Can we make this intuition more
precise? We'd really like to see something more formal.
\end{quote}
We believe that the implicit holistic component of this “hand-wavy” approach is central to the quality of \skipgr{} representations and we can make the intuition more precise by analogy with the definition of the \emph{truth-value potential}.

\begin{figure}
	\centering
	{\scshape
	\tikzset{every picture/.append style={scale=0.6, every node/.style={scale=0.75}}}
		\hbox{}\hfill
\begin{tikzpicture}
[   cnode/.style={draw=black,fill=#1,minimum width=8mm,circle},
    normal arrow/.style={draw,-triangle 45,thick},
    big box/.style={draw, minimum width=10mm, minimum height=12em, rectangle},
    small box/.style={draw, minimum height=7mm, rectangle}
]
	\node (m) at (0,0) {man};
	\node (w) at (3,1) {woman};
	\path[normal arrow, color=blue] (m) -- (w);
	\node (u) at (1,-1) {uncle};
	\node (a) at (4,0) {aunt};
	\path[normal arrow, color=blue] (u) -- (a);
	\node (k) at (0.5,-2.5) {king};
	\node (q) at (3.5,-1.5) {queen};
	\path[normal arrow, color=blue] (k) -- (q);
\end{tikzpicture}\hfill\vrule\hfill
\begin{tikzpicture}
[   cnode/.style={draw=black,fill=#1,minimum width=8mm,circle},
    normal arrow/.style={draw,-triangle 45,thick},
    dashed arrow/.style={draw,-triangle 45,thick, dashed},
    big box/.style={draw, minimum width=10mm, minimum height=12em, rectangle},
    small box/.style={draw, minimum height=7mm, rectangle}
]
	\node (ks) at (-0.5,0) {kings};
	\node (qs) at (2.5,1) {queens};
	\node (k) at (0.5,-2.5) {king};
	\node (q) at (3.5,-1.5) {queen};
	\path[normal arrow, color=blue] (k) -- (q);
	\path[normal arrow, color=red] (k) -- (ks);
	\path[normal arrow, color=red] (q) -- (qs);
\end{tikzpicture}\hfill\hbox{}}
	\caption{Examples of embeddings semantic relations according to \cite{mik}.}
	\label{fig:relations}
\end{figure}
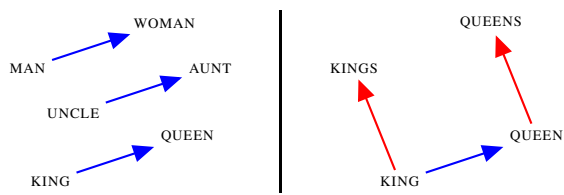
\subsection{Semantic properties of the Skip-Gram model}

The \skipgr{} model was introduced by Tomáš Mikolov et al. \cite{mikolov2013efficient} as a method to
efficiently train word embeddings. It exceeded state-of-the-art in various
semantic tasks. The embeddings have interesting semantic properties, most
notably the vector arithmetic illustrated by Figure~\ref{fig:relations} and the following equation \cite{mik}:
$$v_{king} - v_{man} + v_{woman} \approx v_{queen}.$$
meaning that starting with the word “king”, if we subtract the vector for the word “man” and add the vector for the word “woman”, the nearest vector in the embedding space will be the one that corresponds to the word “queen”. This means that \emph{queen} is to \emph{woman} as \emph{king} is to \emph{man}.

Hollis et al. \cite{Hollis2016} show that it is possible to infer various psycholinguistic and semantic properties of words from the \skipgr{} embeddings.
Mikolov et al. \cite{mikolov2013distributed} also trained the \skipgr{}
model with phrases, resulting in even simpler and more elegant equations, such as
$$v_{Germany} + v_{capital} \approx v_{Berlin}.$$

\begin{figure}
	\centering
	{\tikzset{every picture/.append style={scale=0.55, every node/.style={scale=0.70}}}
	\begin{tikzpicture}
[   cnode/.style={draw=black,fill=#1,minimum width=10mm,circle},
    normal arrow/.style={draw,-triangle 45,very thick},
    dashed arrow/.style={draw,-triangle 45,very thick,dashed},
    small box/.style={draw, minimum height=7mm, minimum width=7mm, rectangle},
    small cross/.style={draw, minimum height=7mm, minimum width=7mm, cross out}
]

	\node[] (p1) at (2,1) {CBOW};
	\node[] (p2) at (10,1) {Skip-gram};

	\node[cnode=white] (s-3) at (2,-3 + 1) {$\sum$};
	\node[small box] (g-3) at (4,-3 + 1) {$w_{3}$};
	\node[small box] (f-3) at (0,-3 + 1) {};
	\node[small cross] (f-3c) at (0,-3 + 1) {};
	\path[dashed arrow] (s-3) -- (g-3);
\foreach \x in {1,2,4,5} {
	\node[small box] (f-\x) at (0,-\x + 1) {$w_{\x}$};
	\path[normal arrow] (f-\x) -- (s-3);
}
	\node[] (f-6) at (0,-6 + 1) {$\vdots$};

	\node[cnode=white] (r-3) at (10,-3 + 1) {};
	\node[small box] (b-3) at (8,-3 + 1) {$w_{3}$};
	\path[normal arrow] (b-3) -- (r-3);
\foreach \x in {1,2,4,5} {
	\node[small box] (n-\x) at (12,-\x + 1) {$w_{\x}$};
	\node[small box] (m-\x) at (8,-\x + 1) {};
	\node[small cross] (mc-\x) at (8,-\x + 1) {};
	\path[dashed arrow] (r-3) -- (n-\x);
}
	\node[] (b-6) at (8,-6 + 1) {$\vdots$};

\end{tikzpicture}}
	\caption{CBOW and \skipgr{} language models according to \cite{mikolov2013efficient}.}
	\label{fig:skipcbow}
\end{figure}
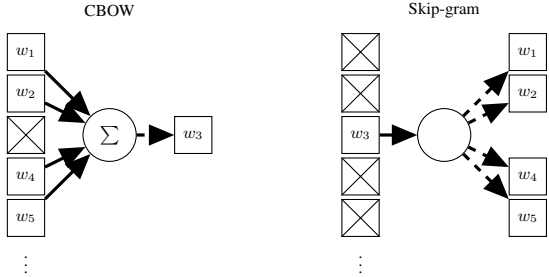

Mikolov et al. \cite{mikolov2013efficient} proposed another shallow neural language model, Continuous Bag of Words (CBOW).
The main difference between CBOW and \skipgr{} (see Figure~\ref{fig:skipcbow}) is that while
\skipgr{} predicts context words from a given word, CBOW predicts a word from a given context.

\section{RELEVANT THEORIES OF MEANING}

In this section, we discuss theories of meaning that are relevant to word representations in artificial neural networks. Notice that even though they strictly speaking do not require meaning holism, they all lean towards it quite strongly.

\subsection{The distributional hypothesis}

Holism is generally a better alternative in cases where there is nothing beside language itself to anchor meaning to. This is the case of neural language (representation) models. If they represent meaning at all, it must be derived from the training corpus. This may be the reason behind the popularity of the distributional hypothesis in neural language model literature. The famous saying by Firth \cite{firth1957}, ``You shall know a word by the company
it keeps!'', is quoted in majority of papers concerned with vector space
models of language. 

The general distributional hypothesis states that
the meaning of a word is given by the contexts in which it occurs.  It is,
however, worth noticing that in Firth's theory, collocation is just one among
multiple levels of meaning and his text does not support the idea of
meaning based on context alone. 

A more suitable formulation of the distributional hypothesis (referenced in connection
to \skipgr{} in \cite{bojanowski2017enriching}) is found in \emph{Distributional
structure} \cite{harris1954distributional}, where it is  suggested that distribution
may be used for comparing meanings and that “difference of meaning correlates
with difference of distribution”.

Although this certainly describes a basic principle of neural language models,
it is still rather vague.

\subsection{The \emph{use theory} of meaning}

The use theory of meaning can be summed up as “the meaning  of a word is its use in the  language” \cite[\S\,43]{witt:pi}. It is associated with late Wittgenstein's concept of language game. 
In \emph{Philosophical Investigations} \cite[\S\S\,499--500]{witt:pi}, he writes:
\begin{quote}
To  say  ``This  combination  of words  makes  no  sense''  excludes it  from
the  sphere  of  language  and  thereby  bounds  the  domain  of language.
[\ldots] When  a  sentence  is  called  senseless,  it  is  not  as  it  were
its sense  that  is  senseless.   But  a  combination  of words  is  being
excluded from  the  language,  withdrawn  from  circulation.
\end{quote}
This “bounding of the domain of language” is precisely what language model does, therefore the use theory may be one way to connect language modelling and semantics.

That “knowledge of language emerges from language use” is also one of main hypotheses of cognitive linguistics  \cite{croft_cognitive_2004}.

\subsection{Structuralism}

In structuralism \cite{de_saussure_course_1916}, the meaning of a word is given by its relation to the other words of the language:
\begin{quote}
The  elements  of  a  structure  have  neither  extrinsic  designation,   nor   intrinsic  signification. Then  what   is   left?  [\ldots{}] [N]othing other than a sense [\ldots{}]: a  sense  which  is  necessarily  and  uniquely  “positional.”
\cite{deleuze1953we}
\end{quote}
This holds for word representations in artificial neural networks as well. The vectors representing the words do not have any other meaning than their position among the rest of the vectors and a single vector does not have any significance outside the model. This is also demonstrated by the vectors being different every time the model is trained because of random initialization.

\section{SKIP-GRAM AND TRUTH-VALUE POTENTIAL}

In this section, we introduce the \emph{truth-value potential} and show that \skipgr{} corresponds to it better than CBOW.

\subsection{The truth-value potential}

Tugendhat's compact reformulation of Frege's sentence holism, the definition of meaning as \emph{truth-value potential} is 
\cite{tug}:
\begin{quote}

[T]wo expressions $\phi$ and $\psi$ have the same truth-value potential if and
only if, whenever each is completed by the same expression to form a sentence,
the two sentences have the same truth-value.

\end{quote}
We can also express this definition in the following form:
$$M(\varphi) = M(\psi) \iff \forall x: T(x(\varphi)) = T(x(\psi)),$$ 
where $M$ is the truth-value potential (meaning), $T$ is the truth-value of the
sentence and $x(\omega)$ is the result of completing the expression $\omega$ by
the expression $x$ to form a sentence.

One important aspect of this definition is that, following Frege \cite{frege1892sinn}, 
it is based on an assumption that the sentence (or rather the corresponding \emph{judgement})
is the basic unit of meaning.

\subsection{Word2vec models and semantic holism}

The definition of meaning as truth-value potential is analogous
to the process of training a model for word representations. One difference is that when we are training a model, we do not have the whole of language at our disposal. 
Even after approximating the language with a finite corpus, it still is not practical to compare all the contexts for a given word at the same time, therefore the universal quantifier has to be  replaced by an iterative process of examining the contexts one by one (or actually batch by batch, which is a step back towards the totality that is being estimated). And we have no means to asses whether the sentences from the corpus are true or false.
We can either assume that they are mostly true, or try to replace the concept of truth with something else (maybe language use). 
Even the first option seems to be enough---imagine a corpus full of false sentences about cats, e.g. “Cats can fly.”, “Cats are cetaceans.” etc. We cannot expect the representation of the word “cats” in a model trained on this corpus to be any good, therefore the requirement for the corpus to consist mostly of true sentences is not excessive.

The simplest model that corresponds to this analogy is the \skipgr{} model. It
does just what is described in the definition -- it fixes a word and goes through
all the possible contexts. It compares the words based on the context. The context words are \emph{predicted} and their representations
are fixed (in a single training step), while the representation of a single
word is \emph{learned}. By learning the representation of a word from the
representation of the context, \skipgr{} complies to the principles of semantic
holism. The analogy
between the definition of truth-value potential and the process of training the \skipgr{} model
is one possible explanation for its semantic properties and its performance in semantic
tasks.

The complementary CBOW architecture (see Figure~\ref{fig:skipcbow}) performs much worse in the evaluation of the semantic tasks \cite{mikolov2013efficient}.  
In CBOW, a missing word is \emph{predicted} from its context. Therefore, in a
single learning step, the representation of the missing word is fixed. What
changes (and is \emph{learned}) is the representation of the context words. By
learning the representation of the context from the representation of the word,
CBOW is implicitly conforming to semantic atomism: words are the basic units of
meaning and the meaning of the broader context is derived from the atomic
meaning of words. This may be the reason why CBOW does not exhibit the same semantic properties as \skipgr{} and it performs worse in semantic tasks.

\section{CONCLUSION AND FUTURE WORK}

The distributional hypothesis as an explanation for the semantic properties
of neural language models should be expanded into a more detailed account. We show
one possible way to do that via a Fregean approach to meaning.

Both the distributional hypothesis itself and Tugendhat's interpretation of
Frege's work are examples of holistic approaches to meaning, where the meaning of the whole determines the meaning of parts. As we demonstrated on the
opposition between \skipgr{} and CBOW models, the distinction between semantic
holism and atomism may play an essential role in semantic properties of neural
language representations models.

We have demonstrated the connection between the \skipgr{} model and the
definition of meaning as truth-value potential. Although this is an isolated
observation of an analogy between a specific model and a specific theory about
meaning, it is a crucial step towards finding a theory of meaning that would
correspond to the current results of NLP research, increasing our understanding
of NLP and ultimately the language itself.

The direction of research from successful language technologies to properties of language itself offers many opportunities for inquiry, with very few being explored so far.

Many state-of-the-art models for natural language processing use smaller units than words for their input and output. This analysis could be extended to take this into account. 

It might also be  interesting to think about the philosophy of science in technical fields dominated by machine learning, but that is far beyond the scope of this paper.

\ack
This work has been supported by the grant 18-02196S of the Czech
Science Foundation.
This research was partially supported by SVV
project number 260 575.

\bibliography{interpret}

\end{document}